\documentclass[a4paper,11pt,fleqn]{article}

\usepackage{geometry} % margins and author names 
\DeclareUnicodeCharacter{2064}{}

\usepackage{setspace} % for doublespacing

\usepackage{amsmath} 
\usepackage{amssymb}
    \usepackage{mathtools}

\usepackage{graphicx}
\usepackage{placeins} % \floatbarrier 
\usepackage{pdflscape} %for making a nice page in landscape format (/lscape)

\usepackage{xcolor}
\definecolor{TsinghuaPurple}{RGB}{79,38,131} % Tsing Hua purple color
\definecolor{ForestGreen}{RGB}{34,139,34}
\definecolor{darkred}{RGB}{190,20,52}

\usepackage{booktabs} % \toprule, \midrule 
\usepackage{rotating} % Rotating table

\usepackage[sort]{natbib} % references in shape [p2]
\usepackage[colorlinks]{hyperref} 
    
\usepackage{comment}
\usepackage{graphicx, subfigure, epsfig, tabularx, threeparttable, tablefootnote, booktabs, stackengine, multirow, multicol, makecell, float}

\usepackage{booktabs} % For professional looking tables
\usepackage{caption} % For customizing captions
\usepackage{lipsum} % For sample text

\usepackage{xurl} % Better URL line breaking
\usepackage{breakurl} % Another option for URL handling

\usepackage{pdfcomment}

\usepackage{siunitx}  % For aligning numbers at decimal points
\newcounter{hypothesiscounter}
\renewcommand{\thehypothesiscounter}{\arabic{hypothesiscounter}}

\newcounter{subhypocounter}

\newcommand{\indep}{\perp \!\!\! \perp}

\definecolor{ForestGreen}{RGB}{34,139,34}

\newcommand{\galit}[1]{\textcolor{black}{#1}}
\newcommand{\david}[1]{\textcolor{black}{#1}}
\newcommand{\travis}[1]{\textcolor{black}{#1}}

\newcolumntype{Y}{>{\centering\arraybackslash}X}
  
\hypersetup{citecolor=blue,linkcolor=blue,urlcolor=blue} % [s1]

\graphicspath{ {./files/}} % standard path for figures
\newcommand{\tablepath}{files} % [a1]
    \makeatletter
    \newcommand{\includetable}[1]{%
      \@ifundefined{tablepath}{%
        \InputIfFileExists{#1}{}{}%
      }{%
        \InputIfFileExists{\tablepath/#1}{}{\InputIfFileExists{#1}{}{}}%
      }
    }
    \makeatother

\begin{document}

\begin{titlepage}
\title{From What Ifs to Insights: \\ Counterfactuals in Causal Inference vs. Explainable AI}

\author{
Galit Shmueli\thanks{Institute of Service Science, National Tsing Hua University, Hsinchu, Taiwan. 
\texttt{\href{mailto:galit.shmueli@iss.nthu.edu.tw}{galit.shmueli@iss.nthu.edu.tw}}}
\and
David Martens\thanks{Faculty of Business and Economics, University of Antwerp, Antwerp, Belgium. 
\texttt{\href{mailto:david.martens@uantwerpen.be}{david.martens@uantwerpen.be}}}
\and
Jaewon Yoo\thanks{Institute of Service Science, National Tsing Hua University, Hsinchu, Taiwan. 
\texttt{\href{mailto:jaewon.yoo@iss.nthu.edu.tw}{jaewon.yoo@iss.nthu.edu.tw}}}
\and
Travis Greene\thanks{Copenhagen Business School, Frederiksberg, Denmark. 
\texttt{\href{mailto:travis.greene@iss.nthu.edu.tw}{travis.greene@iss.nthu.edu.tw}}}
}

\date{\today}
\maketitle
 
\begin{abstract}
\thispagestyle{empty}
\noindent Counterfactuals play a pivotal role in the two distinct data science fields of
causal inference (CI) and explainable artificial intelligence (XAI). While the core idea behind counterfactuals remains the same in both fields---the examination of what would have happened under different circumstances---there are key differences in how they are used and interpreted. We introduce a formal definition that encompasses the multi-faceted concept of the counterfactual in CI and XAI. We then discuss how counterfactuals are used, evaluated, generated, and operationalized in CI vs. XAI, highlighting conceptual and practical differences. By comparing and contrasting the two, we hope to identify opportunities for cross fertilization across CI and XAI.
\\[0.4cm] 

\noindent \textbf{Acknowledgements:} The authors acknowledge the support of the Research Foundation Flanders (FWO), Grant G0G2721N.

\end{abstract}

\end{titlepage}

\newpage

%-------------------------------------------------------------------------------
\section{Introduction}
%-------------------------------------------------------------------------------

The concept of \galit{the \emph{counterfactual}---the opposite of a fact---}has played a pivotal role in \galit{two distinct data science areas: in causal inference (CI) and in explaining predictions (explainable AI, or XAI). While both areas have made important contributions to theory and practice, they focus on different types of modeling~\citep{shmueli2010explain}: CI focuses on causal explanatory modeling, while XAI arises in the context of predictive modeling.} 
\galit{The concept of the counterfactual in CI and XAI} hinges upon the same fundamental premise---\textit{what if} scenarios. Yet, these two \galit{uses} leverage the concept distinctively along multiple dimensions, including purpose, causality vs. correlation, the estimated quantity of interest, aggregation level, the modified object, and assumptions. These differences offer valuable insights into the nuances of counterfactual \galit{uses, highlighting potential avenues for cross fertilization between the distinct areas of CI and XAI. In this work we focus on the \emph{counterfactual-oriented causal inference approach} (also known as the ``potential outcomes framework''), and on the XAI approach of \emph{counterfactual explanations}. We discuss how counterfactuals} are defined, used, evaluated, generated, and operationalized \galit{in CI vs. XAI, highlighting conceptual and practical differences. By comparing and contrasting the two, we hope to identify opportunities for cross fertilization across CI and XAI.}

%-------------------------------------------------------------------------------
\subsection{Counterfactual Definition}
%-------------------------------------------------------------------------------

While the term \emph{counterfactual} is heavily used by CI and XAI researchers, we find a gap in terms of a clear and comprehensive definition. Some authors introduce the counterfactual via an example \citep[e.g., ][]{glymour2016causal,pearl2018book, wachter2017counterfactual}. CI researchers typically use ``counterfactual outcome" to refer to an outcome that did not materialize, as in ``counterfactual (or potential) outcomes'' \citetext{e.g., \citealp{sjolander2012language}; \citealp[p.~254]{pearl2018book}; \citealp{dickerman2020counterfactual}}, while in XAI counterfactual refers to the inputs, such as: 
``[a counterfactual is a] statement of how the world would have to be different for a desirable
outcome to occur.'' \citep{wachter2017counterfactual}. To be able to discuss, compare and contrast counterfactuals in CI and in XAI, we seek a definition that encompasses both the outcome and inputs (treatment). The definition by \citep{glymour2016causal} is a good start:
\begin{quote}
    \ldots an “if” statement in which the “if” portion is untrue or unrealized—is known as a counterfactual. The “if” portion of a counterfactual is called the \emph{hypothetical condition}, or more often, the \emph{antecedent}. 
\end{quote}
We propose the following more comprehensive definition, which specifically refers to outcomes (outputs) and treatments (inputs):
\begin{quote}
    A \emph{counterfactual} is an IF-THEN statement where the IF portion, which describes the treatment group or input values, is untrue or unrealized, and the THEN portion describes the potential outcome.
\end{quote}

%-------------------------------------------------------------------------------
\subsection{A Brief History of the Counterfactual Across Disciplines}
%-------------------------------------------------------------------------------

Counterfactuals have a rich history in philosophy, statistics, econometrics, and computer science. The concept was first formally introduced by philosopher David Lewis in 1973, who described counterfactuals as subjunctive conditionals in the context of ``possible worlds.'' In Lewis's semantics, a counterfactual statement like ``If A were the case, then B would be the case'' is true if and only if B is true in the closest possible world in which A is true \citep{lewis1973counterfactuals}. This concept is a cornerstone of philosophical debates, particularly in the philosophy of mind and metaphysics.

In 1974, statistician Donald Rubin introduced the ``potential outcomes'' framework, which is fundamentally a counterfactual interpretation of causality. In Rubin's framework, the causal effect of an intervention is defined as the difference between the potential outcomes with and without the intervention \citep{rubin1974estimating}. This interpretation has been influential in econometrics, biostatistics, and social science in general, providing a basis for causal inference from observational data. \galit{The counterfactual is also a key component of computer scientist Judea Pearl's structural causal modeling (SCM) approach for identifying causal effects, where counterfactuals are derived from a functional causal model \cite[][p. 35]{pearl2009causality}. The SCM counterfactual reasoning provides machinery for answering counterfactual questions from deterministic and probabilistic models \cite[][p. 201]{pearl2009causality}.}

\galit{In the context of prediction, especially from ``black box'' models, \emph{counterfactual explanations} provide} \david{an explanation for the predicted outcome of a single instance.} \galit{The counterfactual explanation aims to} explain why a classifier made a certain prediction. \galit{This predicted class  typically leads to a decision regarding the predicted entity (e.g., accept/reject for a loan seeker).} 

\david{The counterfactual explanation was first proposed in the context of document classification for explaining %document classifications, and more specifically: 
why a predictive model labeled a website as containing adult content or not~\citep{martens2014explaining}. Insight into the predicted class is provided by describing the words that, had they not appeared on the website, would change the predicted class. This was later generalized to tabular data \citep{wachter2017counterfactual}. In this setting, a counterfactual explains what caused the predictive model to label a certain unit as belonging to a certain class in terms of \emph{changes in the feature-value} combinations that would lead to a different \emph{predicted outcome}. Note that other popular XAI methods, such as LIME and SHAP, that explain a \galit{numerical} prediction score rather than a \galit{categorical predicted} outcome \galit{(predicted class)}, do not provide a counterfactual (see~\citealp{fernandez2020explaining} for a more detailed discussion.)}

%-------------------------------------------------------------------------------
\subsection{Causal Inference and Counterfactual Explanations in Practice}
%-------------------------------------------------------------------------------

Empirical research \travis{in numerous disciplines including} marketing, information systems (IS), and strategic management, frequently employs the concept of counterfactuals primarily within the realm of CI \citep{goldfarb2011search, flammer2016impact, tirunillai2017does, xu2017battle}. Predominantly, these studies \travis{focus} on identifying causal relationships, such as the effect of advertising on purchase conversion, using a mix of experimental and observational methodologies. \galit{Similarly,} \travis{counterfactual queries are common in public health and medicine---two fields frequently faced with urgent, high-stakes questions about the causal effects of various drugs, behavioral interventions, therapies, and surgical procedures. Medical researchers and practitioners wish to know the answers to practically and theoretically important counterfactual questions; they often turn to clinical trials as well as observational studies to answer such questions.} \galit{Using our counterfactual definition, the focus in causal inference is on the THEN outcomes that result from pre-selected IF scenarios.}
While \galit{causal inference} methods have proved instrumental in establishing causality, there is a growing opportunity to further leverage counterfactual explanations, particularly in data-rich environments, to garner additional insights and actionable implications.

\emph{Counterfactual explanations}, a concept extensively used in XAI, can help researchers and practitioners understand how changes in the input variables could lead to different outcomes, as predicted by the AI model. \galit{Returning to our definition of a counterfactual, in counterfactual explanations the focus is on identifying the IF scenario that leads to a change in the THEN portion.} While CI primarily deals with \travis{estimating} the overall effect of a treatment or an intervention (e.g., advertising) on an outcome (e.g., purchase conversion) \travis{in aggregate}, counterfactual explanations \travis{provide prescriptions for action by} exploring how different aspects of the treatment \galit{and input values} could lead to different (predicted) outcomes, generally \travis{for specific individuals.} \travis{Merging methods and concepts from CI and XAI may thus provide useful bridge between descriptive, predictive, and prescriptive applications of data science.}

Consider the example of studying the causal effect of digital advertising on purchase conversion. Traditional CI might estimate the average effect of advertising on purchase conversion.\footnote{The customer's behavior of making an actual purchase of the advertised product--the advertising industry commonly refers to these as ``purchase conversions'' \citep{gordon2019comparison}.} But this leaves unanswered questions such as:
\begin{itemize}
    \item How would a change in the content or placement of \galit{a specific}    advertisement have affected the conversion rate?
    \item How would a different advertising strategy have performed with a particular demographic?
\end{itemize}
Counterfactual explanations can \david{assist in} providing answers to these questions, offering nuanced insights into the complex relationships between features, interventions, and outcomes. \travis{One example of a counterfactual explanation used in the domain of advertising leverages a learned classifier to suggest feature transformations that, if realized, \galit{would} turn a \galit{predicted} low quality advertisement into a high quality \galit{prediction} \citep{tolomei2017interpretable}. A hypothetical explanation might be something like: \emph{if your ad's text readability score \galit{increased from medium to high}, then your ad's predicted class would change from low to high quality.}} 

\galit{Another example that clearly contrasts the types of counterfactual queries in CI and XAI is job recruitment. Here are two counterfactual queries corresponding to CI vs. XAI: }
\begin{itemize}
    \item (CI): \galit{Does race affect the probability of being invited to a job interview?}
    \item (XAI): If your resume would have included Python and AI then your prediction would change from rejected to invited for an interview.
\end{itemize}

In the context of criminal justice, the recent controversy over the use of algorithmic risk assessment tools\footnote{\url{https://www.propublica.org/article/machine-bias-risk-assessments-in-criminal-sentencing}} again highlights differences in CI and XAI counterfactual queries. For example:
\begin{itemize}
    \item (CI): Is the algorithm biased against African-Americans?
    \item (XAI): Had your age been older than 30 and race=Caucasian then your prediction would change from high-risk to low-risk.
\end{itemize}

\galit{These examples are meant to give an initial glimpse of how the counterfactual plays different roles in CI vs.~XAI.}

%-------------------------------------------------------------------------------
\subsection{\galit{When Worlds Collide}}
%-------------------------------------------------------------------------------

\galit{While the counterfactual has proven an exceptionally useful concept in both causal inference and XAI, to the best of our knowledge these developments have mostly been occurring in parallel without considering how they might differ, and importantly, how they can benefit each other: How might causal explanation benefit from counterfactual predictions? Likewise, how might counterfactual explanations make use of \travis{CI} counterfactuals?} 

Secondly, empirical research has traditionally focused on causal inference. Yet, there is significant potential in incorporating the use of counterfactual explanations to enrich the theoretical as well as managerial implications of causal empirical research. \galit{By uncovering the underlying differences between CI and XAI, we hope that new research and applications will emerge that combine CI and XAI in order to} gain deeper insights, make more informed decisions, and provide more actionable recommendations.

%-------------------------------------------------------------------------------
\subsection{Terminology and Notation}
%-------------------------------------------------------------------------------

\begin{figure}[h]
    \begin{center}
    \caption{Illustration of Counterfactuals in CI vs.~XAI}
    \includegraphics[scale = .8]{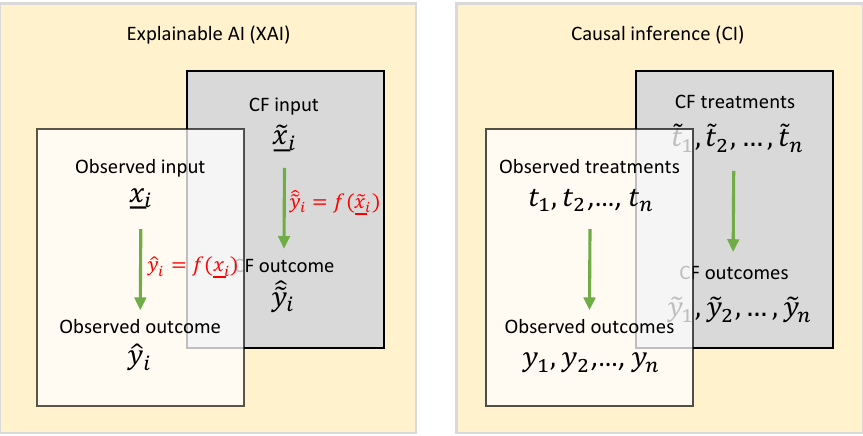}
    \label{fig:schematic}
    \end{center}
\end{figure}

The fields of \travis{CI} and \travis{XAI} use different terms and notation. For simplicity, we use the following:
\begin{description}
    \item[Unit ($i$)] refers to a single observation, record, or instance. For a unit we can measure input values, an outcome, generate a predicted outcome, etc.\footnote{For example, in a study on customer purchasing behavior, a unit could be a single customer transaction, where the input values might include the customer's demographics and the items purchased, the outcome could be purchase frequency, and a predicted outcome might be the estimated purchase frequency based on a predictive model.} % Jaewon: please let me know if I am going too crazy with the footnotes!
    \item[Intervention/treatment (group) ($T$)] refers to the act of externally manipulating a variable in the system to take on a certain value ($T=t$) irrespective of its original state.\footnote{This is fundamentally different from merely observing a variable taking on a certain value naturally. Thus, in an observational setting, the concept extends to finding ways to ``induce'' a hypothetical intervention, often by leveraging exogenous variations in treatments.} %[JAEWON PLS HELP]
    \item[Features ($\underline{x}_i$)] refer to values of the input variables ($\underline{X}_i$) measured for unit $i$. For example, features for a website user might include the user's demographics and past behavior on the website. 
    \item[Outcome ($y_i$)] refers to a measured value for unit $i$.
    \item[Potential outcome ($\tilde{y}_i$)] refers to an unobservable quantity for unit $i$ that would theoretically be possible to measure had the unit been \travis{assigned} to a different treatment group. 
    \item[Prediction, or predicted outcome, or predicted class ($\hat{y}_i$)] refers to the class %a value generated 
    assigned to unit $i$ by a predictive model, given the unit's features $\underline{x}_i$.
\end{description}
Figure \ref{fig:schematic} provides a schematic of the different terms and notations.

%-------------------------------------------------------------------------------
\section{Counterfactuals in CI vs. XAI}
%-------------------------------------------------------------------------------

\galit{In this section we examine key dimensions that distinguish between the CI counterfactual and the XAI counterfactual. These include the purpose of using the counterfactual, what is modified, what is estimated/computed, the aggregation level, and assumptions, \david{as summarized by Table~\ref{taboverview}}.}

\begin{table*}[htbp]
\centering
\caption{\david{Overview of differences in use of counterfactuals}}
\label{taboverview}
\begin{tabular}{|l|l|l|}
\hline
\textbf{Dimension} & \textbf{Causal Inference} & \textbf{Explainable AI} \\
\hline
\hline
Purpose & Estimate causal effect & Explain a predicted outcome \\
Type of Relationship & Causal & Correlation \\
Quantity of Interest & Difference in outcomes & Input changes \\
Aggregation Level & Sample & Unit \\
Modified Object & Model & Data \\
Assumptions & On data generating function & On given predictive model \\
Performance Evaluation & Standard error, conf. interval, %RDD, McCrary tests, etc. 
& Feasibility, sparseness, etc. \\
& significance, robustness tests  & \\
\hline
\end{tabular}
\end{table*}

%-------------------------------------------------------------------------------
\subsection{Purpose: \galit{Estimate a Causal Effect vs. Explain a Predicted Outcome}}
%-------------------------------------------------------------------------------

Counterfactuals serve distinct but interrelated purposes in CI and XAI. In both domains, they fundamentally inform decision-making, albeit with different emphases and \travis{in different contexts and for different audiences.}

In the realm of CI, which has its roots in applied econometrics \galit{and statistics}, counterfactuals are primarily deployed to \galit{estimate the causal effect of an intervention ($T$) on an outcome of interest ($Y$). Ideally, we would want to compare the effect of the intervention ($T=1$) to that of no intervention ($T=0$ \galit{or alternative interventions), such as in the case of evaluating a new policy or medication}. The notion of the counterfactual is used to} tackle the fundamental problem of causal inference \citep[FOCI; ][]{holland1986statistics}, \galit{which is} %. The essence of FOCI is 
the challenge of simultaneously observing all %potential 
outcomes for a given unit $i$ \galit{under each intervention/non-intervention}. This gives rise to the missing data problem: if we observe \galit{the} outcome \galit{($y_i$) for unit $i$ who was given the intervention} ($t_i=1$), what would be the outcome $\tilde{y}_i$ \galit{without intervention} ($\tilde{t}_i=0$)? \galit{Or vice versa: if we observe the outcome under no intervention, what would be the outcome under intervention? These missing outcomes are called \emph{potential outcomes}.} Various identification strategies have been proposed to circumvent this issue which essentially revolve around discerning the effect of an intervention or treatment under a hypothetical scenario, informing the decision of whether or not to implement a treatment. 

\galit{In contrast,} the field of XAI \david{uses} counterfactuals primarily to elucidate the inner workings of complex (often referred to as ``black box'') \galit{predictive} models. \galit{Specifically, a \emph{counterfactual explanation} \citep{martens2014explaining,wachter2017counterfactual,guidotti2022counterfactual} describes how the ``nearest possible world'' ($\tilde{\underline{x}}_i$) has to be different for a desirable outcome ($\hat{\tilde{y}}_i$) to occur \citep{wachter2017counterfactual}. \travis{Counterfactual explanations can help data scientists ``debug'' predictive models by detecting spurious correlations and unwanted  biases in the training data. But arguably their most valuable application is in helping individuals receiving a negative decision from an algorithmic predictive model---such as a loan rejection---understand what would have been needed to receive a positive decision.} Here, the focus is not on uncovering causal relationships per se, but rather on augmenting the interpretability/comprehensibility and transparency of the \galit{predictive} model, describing what causes \textit{the model} ($f$) to generate a certain prediction. The primary question asked might be: ``\galit{For instance $i$,} which set of features should we change, and by how much ($\tilde{\underline{x}}_i$), to achieve a certain model output ($\hat{\tilde{y}}_i$)?'' \travis{The answer to this question can} empower users to understand the decision-making process of the model and to make informed decisions based on these insights \citep{venkatasubramanian2020philosophical}.}

%-------------------------------------------------------------------------------
\subsection{Causality vs. Correlation}
%-------------------------------------------------------------------------------

The contrast between %these two applications 
\galit{the use} of counterfactuals \galit{in CI vs. XAI} underscores an important distinction: While CI aims to comprehend \galit{and estimate} %the 
\emph{causal} effects \galit{between interventions ($t$) and measurable outcomes ($y$) in the real world, counterfactual explanations in XAI are based on \emph{correlations or associations} between the predicted outcome ($\hat{y}$) and the input variables ($\underline{x}$) \travis{as found in the training data}. The counterfactual explanation is based on the correlation structure of the black box predictive model ($f$), linking the predictions and inputs of unit $i$.} \travis{There is no guarantee, however, that observed correlations in the training data accurately reflect causal structures (i.e., data generating processes) operating in reality. Unlike CI-based empirical research, where typically a domain theory guides reasoning about the underlying causal mechanisms and relations underlying the observed data \citep{hofman2021integrating}, in XAI it is not obvious what an analogous ``domain theory'' would look like for a model designed, for example, to predict loan default.\footnote{In some cases, however, legal restrictions might provide a basis for including or excluding model features related to race or gender.} For this reason, the predictive models used in XAI can become decoupled from reality and lead to dubious recommendations, e.g. that individuals make causally ineffective feature changes \citep{karimi2022survey}.}

%-------------------------------------------------------------------------------
\subsection{Quantity of interest: Treatment Effect vs. Feature Values}
%-------------------------------------------------------------------------------

The quantity of interest in CI is a causal estimand, whereas in the XAI counterfactual explanation interest is in a hypothetical feature value combination. The quantities generated for these purposes end up being different in terms of outcomes vs.~inputs as well as in terms of level of aggregation. We explain each next.

\subsubsection{Estimated Quantities: Outcomes vs. Inputs}
In CI the main quantity of interest, or \emph{estimand}, is the \emph{difference between two outcomes}. Ideally, for each unit $i$ we would want $y_i-\tilde{y}_i$ (the difference between the measured outcome and the counterfactual outcome\footnote{A causal effect is a comparison of potential outcomes, for the same unit, at the same moment in time, post treatment \cite[][p. 6]{imbens2015causal}}). Decisions by policymakers would then be based on this (ideal) quantity of interest. 
In contrast, in XAI, the quantity of interest is the ``best" combination of feature values for unit $i$: $\tilde{\underline{x}}_i$, which we might call \emph{counterfactual input values}. Decisions (by individual $i$) would then be based on $\tilde{\underline{x}}_i$.
In other words, in CI the quantity of interest is the THEN portion of the counterfactual definition, whereas in XAI it is the IF portion.

\subsubsection{Aggregation Level: Sample vs. Individual Unit}
As described above, counterfactual explanations remain at the individual level ($i$). However, while the ideal causal effect in CI is at the individual level, due to the FOCI, we cannot estimate the individual-level quantity of interest $y_i-\tilde{y}_i$. Instead, CI estimation resorts to group averages that rely only on observed outcomes ($y_i, i=1,\ldots n$). Examples include the Average Treatment Effect (ATE) that computes the difference between the average outcome of the treated and the average outcome of the untreated; and subgroup averages such as the Average Treatment Effect on the Treated (ATT), or Conditional Average Treatment Effect (CATE). More intricate averages are used when we have a mediator and want to study the mediation mechanism, or when we have multiple treatment variables \citep{acharya2016explaining, imai2011unpacking}. In all these cases, estimands are based on averages, and therefore CI formulas are typically written using expected values (see Table \ref{tab:ATE}).

\begin{table*}[]
    \caption{Common estimands of causal effects in CI}
    \label{tab:ATE}
    \centering
    \begin{tabular}{|lll|}
    \hline
    Acronym & Description & Formula \\ \hline
    \hline
    ATE & average treatment effect     & $\mathbb{E}[Y_{i}(1)-Y_{i}(0)]$ \\
    ATT & average treatment effect on the treated & $\mathbb{E}[Y_{i}(1)-Y_{i}(0)|T_{i}=1]$ \\
    CATE & conditional average treatment effect  & $\mathbb{E}[Y_{i}(1)-Y_{i}(0)|\textbf{X}_{i}=x]$ \\
    ACDE & average controlled direct effect  & $\mathbb{E}[Y_{i}(1,t_{2})-Y_{i}(0,t_{2})]$ \\
    ACME & average causal mediation effect & $\mathbb{E}[Y_{i}(t,M_{i}(1))-Y_{i}(t,M_{i}(0))]$ \\
    ANDE & average natural direct effect  & $\mathbb{E}[Y_{i}(1,M_{i}(t))-Y_{i}(0,M_{i}(t))]$ \\  \hline
    \end{tabular}
\end{table*}

While the above aggregation levels in XAI and CI are most common, we note that aggregation can also be on a continuum: counterfactual explanations can be aggregated across multiple units to learn more about the data and the model, and CI estimands try to get closer to individual level effects, e.g. by using ``synthethic controls" \citep{abadie2003economic}.

%-------------------------------------------------------------------------------
\subsection{What is Modified? Model vs. Data} %Target of Modification}
%-------------------------------------------------------------------------------

A further distinction lies in the targets of modification in these fields. In CI, the target of modification often involves changing the structural form of the model or adjusting the treatment variable in an attempt to simulate different hypothetical scenarios. This is consistent with the purpose of CI, which is to estimate the causal effect of a treatment or intervention on an outcome. In this context, counterfactuals are used to represent the outcome that would have occurred had the treatment been different. This often requires careful model selection and testing to ensure that the counterfactual scenarios are plausible and consistent across different models.

In contrast, in the field of XAI, the target of modification is the input data to a given, fixed predictive model. Counterfactuals in XAI are used to illustrate how changes in the input features would have led to different model output. Rather than changing the model itself, XAI focuses on understanding how the model responds to changes in the input data. This approach is useful for interpreting complex models, as it provides insights into how important each feature is and, in particular, how much these features should change to get a different model output (counterfactual explanation).

%-------------------------------------------------------------------------------
\subsection{Assumptions} \label{sec:assumptions}
%-------------------------------------------------------------------------------

The fields differ in the assumptions they make and the emphasis they place on understanding the system they are working with. CI relies on a multitude of assumptions that represent beliefs or assertions about the underlying structure and processes of the system that cannot be directly tested with the observed data. Consequently, significant emphasis is placed on understanding the data generation process (DGP). 
Underlying Rubin's potential outcomes framework are several well known assumptions that enable identification and estimation of causal effects:
\begin{enumerate}
    \item \textit{Consistency}: %$Y_{i}=Y_{i}(d) \text{ if } D_{i}=d$ \\
        For each unit, there are no different versions of each treatment level. This allows us to connect the counterfactual outcomes with the observed data. %(i.e., $Y_{i} = D_{i} \cdot Y_{i}(1) + (1-D_{i}) \cdot Y_{i}(0)$).
    \item \textit{No interference between units}\footnote{Assumptions (1) and (2) are collectively referred to as stable unit-treatment variation assumption (SUTVA).}: % $Y_{i}(D_{1}, D_{2}, ..., D_{N})=Y_{i}(D_{i})$ \\
        The treatment of one unit does not affect the outcomes of other units.
    \item \textit{Conditional unconfoundedness}\footnote{Also referred to as \textit{weak ignorability}, \textit{selection on observables}, \textit{conditional exchangibility}, or \textit{no unmeasured confounding}.}: %$\{Y_{i}(1), Y_{i}(0)\} \indep D_{i}|\textbf{X}_{i}$ \\
        Conditional on a vector of covariates (i.e., confounders), treatment assignment does not depend on any potential outcomes.
    \item \textit{Positivity/Overlap}: %$0<\mathbb{P}[D_{i}=1|\mathbf{X}_{i}]<1$ \\
        Treatment and control are both possible at every level of $\mathbf{X}_{i}$ (common support).
\end{enumerate}

These are strong assumptions that may not hold in practice. Specifically, assumptions (3) and (4) are based on the premise that the researcher has complete knowledge of the causal structure between the treatment and the outcome. Thus, many of the widely used CI methods suggest relaxing these assumptions and instead proposing a set of more plausible assumptions (see Appendix A). 

\galit{In contrast, the assumptions of counterfactual explanation methods concern the predictive model. Specifically:
\begin{enumerate}
    \item The predictive model is given, and is deterministic. In other words, we can apply the predictive model to score any feature vector of interest ($\underline{x}_i$), and the resulting predicted class ($\hat{y}_i$) will be reproducible.
    \item Some counterfactual explanation methods require access to the predictive model's training data, while others require access to some aspect of the predictive model itself (e.g. its gradients).
\end{enumerate}
}

%-------------------------------------------------------------------------------
\subsection{Performance Evaluation}
%-------------------------------------------------------------------------------

\galit{Because the end goal in CI and XAI is different, namely to estimate a causal effect (CI) and to provide a counterfactual explanation (XAI), measuring the quality of the resulting object differs. In CI we are interested in evaluating the quality of the estimate (e.g. ATE), whereas in XAI we are interested in evaluating the quality of the explanation. The measures in both cases are significantly different.}
In CI, counterfactuals are used to ensure the reliability of the causal estimates and to validate the assumptions underlying the CI methods. This is typically achieved %with a variance estimator to compute 
\galit{by estimating} the standard error and constructing confidence intervals. \galit{In addition, to test the validity of the identification assumptions, a series of robustness checks is conducted}. For instance, in Regression Discontinuity Design (RDD), robustness checks such as the McCrary test and covariate smoothness test are employed \citep{dickerman2020counterfactual,flammer2015does}.\footnote{Similarly, in the difference in differences approach, data scientists typically present empirical evidence of parallel trends between the treated and control groups \citep{xu2017battle}.} These tests are designed to ensure that there is no manipulation around the cutoff and that the covariates are smooth around the threshold, respectively. These robustness checks serve as performance indicators for the CI methods, helping to ensure the reliability of the causal estimates.

On the other hand, in XAI, counterfactuals are used differently, and their performance is evaluated based on a variety of metrics. These metrics are designed to assess the quality of the counterfactual explanations generated by the XAI model. For instance, a variety of performance measures are considered, \galit{including how realistic the counterfactual modifications are (\emph{feasibility}), how close the counterfactual is to the instance's input data (\emph{proximity}), how short the explanation is (\emph{sparsity}), and how robust the counterfactual is to model changes (\emph{stability})~\citep{guidotti2022counterfactual,karimi2022survey,verma2020counterfactual, sokol2020one}}.

%-------------------------------------------------------------------------------
\section{Concluding Remarks}
%-------------------------------------------------------------------------------

\travis{The analysis we  presented clarifies differences in the use of counterfactuals in both causal inference and in explainable AI. Our discussion centered on how counterfactuals
are defined, used, evaluated, generated, and operationalized in these two domains. We hope our discussion shines light on several possible areas of synergy between the fields of causal inference and explanainable AI. Towards this end, we briefly mention a few areas of research that combine prediction with counterfactual explanations.} 

\travis{One salient example of fusing prediction with causal modeling comes from a stream of machine learning research studying issues of \emph{counterfactual fairness} \citep{kusner2017counterfactual}. Predictive counterfactual fairness techniques \citep{goethals2023precof} could help to identify implicit biases in a learned model by revealing counterfactual explanations such as \emph{if you were not a woman, you would have received the loan}. These counterfactual explanations can stimulate further research into the social and institutional mechanisms generating the data used to train the predictive model. Conversely, causal models can help ensure that returned counterfactual explanations are truly actionable and therefore useful to end users.}

\travis{Another interesting use of the counterfactual concept is in evaluating, predicting, and explaining the effects of personalized interventions in both business \citep{schnabel2016recommendations, bottou2013counterfactual} and high-stakes medical contexts \citep{derubeis2014personalized, oberst2019counterfactual}. For instance, there is growing interest in predicting the counterfactual effects of individual healthcare interventions by leveraging causal domain knowledge \citep{prosperi2020causal}. In some AI-based clinical decision-making systems \citep{komorowski2018artificial}, the counterfactual of interest involves estimating how a \emph{policy} (i.e., a decision rule for assigning treatments or recommendations on the basis of observed covariates) would have performed in an online experiment if it had been used instead of the (human) policy that actually generated the data \citep{joachims2021recommendations}. One might also want to know \emph{why} a policy performs better than the current one. To answer this kind of causal question, one technique counterfactually removes individual observations from medical intervention datasets and evaluates the resulting impact of these observations on the value of a proposed policy \citep{gottesman2020interpretable}. These influential data points can then be presented to experts for closer analysis, ultimately fostering potential insights into how to better sequence medical interventions for a variety of patient populations and conditions.}

%A similar strand of counterfactual-based research is found in both recommender systems and

\travis{In summary, the counterfactual is an increasingly popular and powerful concept for imagining novel answers to \emph{what if} scenarios commonly encountered in science, medicine and public policy, business and beyond. With greater appreciation of the conceptual and methodological subtleties of counterfactual analysis, we see opportunities to both improve our understanding of the causal mechanisms in a given domain and explain the predictions of complex models.}

\newpage
\singlespacing
\bibliographystyle{apalike}
\bibliography{references}

\newpage

\appendix % replaced section number to A
\setcounter{table}{0}
\setcounter{figure}{0}

\renewcommand{\thefigure}{A.\arabic{figure}} %changes the \begin{figure}
\renewcommand{\thetable}{A.\arabic{table}} 

\section{Appendix}

\subsection{Identification Assumptions in CI}
Depending on the study design, there might be further assumptions in addition to the four assumptions stated in Section \ref{sec:assumptions}.

\galit{For simplicity of exposition, we use} Rubin's \galit{potential outcomes} causal model notation for stating these assumptions:
 Consider an observed random treatment assignment ($D_{i}$), observed- and potential outcomes ($Y_{i}$ and $Y_{i}(d)$), an instrument ($Z_{i}$), a running/forcing variable ($R_{i}$),  group status (treated or control) ($G_{i}$), a treatment exposure/uptake ($W_{i}$), and a vector of pretreatment covariates ($\textbf{X}_{i}$) \citep{rubin1974estimating}.

The assumptions from Section \ref{sec:assumptions} can be written as:
\begin{enumerate}
    \item \textit{Consistency}: $Y_{i}=Y_{i}(d) \text{ if } D_{i}=d$ \\
        For each unit, there is no different versions of each treatment level. This allows us to connect the counterfactual outcomes with the observed data (i.e., $Y_{i} = D_{i} \cdot Y_{i}(1) + (1-D_{i}) \cdot Y_{i}(0)$).
    \item \textit{No interference between units}: $Y_{i}(D_{1}, D_{2}, ..., D_{N})=Y_{i}(D_{i})$ \\
        The treatment of one unit does not affect the outcomes of other units.
    \item \textit{Conditional unconfoundedness}: $\{Y_{i}(1), Y_{i}(0)\} \indep D_{i}|\textbf{X}_{i}$ \\
        Conditional on a vector of covariates (i.e., confounders), treatment assignment does not depend on any potential outcomes.
    \item \textit{Positivity/Overlap}: $0<\mathbb{P}[D_{i}=1|\mathbf{X}_{i}]<1$ \\
        Treatment and control are both possible at every level of $\mathbf{X}_{i}$ (common support).
\end{enumerate}

Further assumptions might be needed for other designs.
For example, instrumental variables (IV) rely on a set of identification assumptions:

\begin{enumerate}
    \item \textit{Exogeneity}: $\mathbb{E}[\varepsilon_{i}|Z_{i}]=0$ \\
        Instrument $Z_{i}$ is a variable that is determined outside of the model or the system being studied. In a randomized controlled trial (RCT) with noncompliance\footnote{Even when we have access to experimental data where units are randomly assigned to receive treatment, we may have people choosing not to comply due to the fundamental human right; i.e., freedom of choice. This means treatment assignment, $D_{i}$, does not guarantee treatment exposure/uptake, $W_{i}$, such that $D_{i} \neq W_{i}$ for all $i$ which introduces some selection bias.} where the instrument is treatment assignment, it can also be interpreted as \textit{randomization} of the instrument.
    \item \textit{Relevance}: $cov(D_{i}, Z_{i}) \neq 0$ \\
        The instrument $Z_{i}$ is `sufficiently'\footnote{Generally, an F-test on instruments ($H_{o}: \; \gamma =0 \text{ where } \gamma$ is the coefficient for the instrument in the $2^{nd}$ stage) is used to compute the F-statistic which is compared to a cutoff value of 12 \citep{stock2002testing}. Recent work argues for a tighter standard, i.e., F-stat $\geq 104.7$ \citep{lee2022valid}.} correlated with the treatment $D_{i}$. In a RCT with noncompliance, the assumption can be interpreted as having one or more units who complies with the treatment (i.e., share of compliers $\neq 0$).
    \item \textit{Exclusion restriction}: $Y_{i}(z,d) = Y_{i}(z',d) \text{ for all }z, \; z', \text{ and } d.$ \\
        Instrument $Z_{i}$ affects outcome $Y_{i}$ only through treatment $D_{i}$. In an RCT, in contrast to an observational setting, this is an assumption that is fairly easy to justify (randomization affects outcome only through exposure to treatment).
    \item \textit{Monotonicity}: $D_{i}(z) \geq D_{i}(z') \text{ for all } z \text{ and } z'$. \\
         Instrument $Z_{i}$ consistently influences the treatment $D_{i}$ in the same direction for all units. In an RCT with noncompliance where we have units of different compliance types, the assumption implies that we do not have units who defy their treatment status (i.e., defiers: $D_{i}(1)=0 \text{ and } D_{i}(0)=1)$.
\end{enumerate}

Other widely used CI methods such as regression discontinuity design (RDD), difference in differences (DiD) or synthetic control (SC) method rely on an identification assumption that is relatively easier to justify (often with empirical evidence). Consequently, \travis{we have seen a surge of empirical work} in a variety of fields, e.g., information systems, marketing, and strategic management using these identification strategies \citep{goldfarb2011search, flammer2016impact, tirunillai2017does, xu2017battle}.

\begin{enumerate}
    \item Parallel trends: $\mathbb{E}[Y_{i1}(0)-Y_{i0}(0)|G_{i}=0] = \mathbb{E}[Y_{i1}(0)-Y_{i0}(0)|G_{i}=1]$ \\
        The secular trends in the control group is a good proxy for how the treated group would have behaved over time in the absence of treatment.
    \item Continuity: $\lim_{x \downarrow c} \mathbb{E}[Y_{i}(d)|R_{i}=c] = \lim_{x \uparrow c} \mathbb{E}[Y_{i}(d)|R_{i}=c] \text{ for all } d$. \\
        The conditional expectation function of the potential outcomes is continuous especially around a given threshold or cutoff ($c$).
\end{enumerate}
\end{document}